\documentclass[pdflatex,sn-mathphys-num]{sn-jnl}% Math and Physical Sciences Numbered Reference Style
\usepackage{todonotes}
\usepackage{fancyhdr}

\usepackage{graphicx}%
\usepackage{multirow}%
\usepackage{amsmath,amssymb,amsfonts}%
\usepackage{amsthm}%
\usepackage{mathrsfs}%
\usepackage[title]{appendix}%
\usepackage{xcolor}%
\usepackage{textcomp}%
\usepackage{manyfoot}%
\usepackage{booktabs}%
\usepackage{algorithm}%
\usepackage{algorithmicx}%
\usepackage{algpseudocode}%
\usepackage{listings}%
\theoremstyle{thmstyleone}%
\theoremstyle{thmstyletwo}%

\theoremstyle{thmstylethree}%

\begin{document}

% \title{From Evidence to Benchmark: Automated Construction and Rubric-based Evaluation of Clinical AI System}
\title{GAPS: A Clinically Grounded, Automated Benchmark for Evaluating AI Clinicians}

%% \author[1,2]{\fnm{Joergen W.} \spfx{van der} \sur{Ploeg} 

% ===== 蚂蚁集团 =====
\author[1,2,3,4,5]{\fnm{Xiuyuan} \sur{Chen}}\email{dr\_chenxy@pku.edu.cn}
\equalcont{These authors contributed equally to this work.}

\author[6]{\fnm{Tao} \sur{Sun}}\email{suntao.sun@antgroup.com}
\equalcont{These authors contributed equally to this work.}

\author[6]{\fnm{Dexin} \sur{Su}}\email{dexin.sdx@antgroup.com}
\equalcont{These authors contributed equally to this work.}

\author[6]{\fnm{Ailing} \sur{Yu}}\email{yuailing.yal@antgroup.com}
\equalcont{These authors contributed equally to this work.}

\author[6,7]{\fnm{Junwei} \sur{Liu}}\email{baishun@antgroup.com}
\equalcont{These authors contributed equally to this work.}

\author[6]{\fnm{Zhe} \sur{Chen}}
\author[6]{\fnm{Gangzeng} \sur{Jin}}
\author[6]{\fnm{Xin} \sur{Wang}}
\author[6]{\fnm{Jingnan} \sur{Liu}}
\author[6]{\fnm{Hansong} \sur{Xiao}}
\author[6]{\fnm{Hualei} \sur{Zhou}}
\author[6]{\fnm{Dongjie} \sur{Tao}}
\author[6]{\fnm{Chunxiao} \sur{Guo}}
\author[6]{\fnm{Minghui} \sur{Yang}}
\author[6]{\fnm{Yuan} \sur{Xia}}
\author[6]{\fnm{Jing} \sur{Zhao}}
\author[6]{\fnm{Qianrui} \sur{Fan}}
\author[6]{\fnm{Yanyun} \sur{Wang}}
\author[6]{\fnm{Shuai} \sur{Zhen}}

\author[1,2,3,4,5]{\fnm{Kezhong} \sur{Chen}}
\author[1,2,3,4,5]{\fnm{Jun} \sur{Wang}}
\author[1,2,3,4,5]{\fnm{Zewen} \sur{Sun}}
\author[1,2,3,4,5]{\fnm{Heng} \sur{Zhao}}
\author[1,2,3,4,5]{\fnm{Tian} \sur{Guan}}
\author[1,2,3,4,5]{\fnm{Shaodong} \sur{Wang}}
\author[1,2,3,4,5]{\fnm{Geyun} \sur{Chang}}
\author[1,2,3,4,5]{\fnm{Jiaming} \sur{Deng}}
\author[1,2,3,4,5]{\fnm{Hongchengcheng} \sur{Chen}}
\author[1,2,3,4,5]{\fnm{Kexin} \sur{Feng}}
\author[1,2,3,4,5]{\fnm{Ruzhen} \sur{Li}}
\author[1,2,3,4,5]{\fnm{Jiayi} \sur{Geng}}
\author[1,2,3,4,5]{\fnm{Changtai} \sur{Zhao}}
\author[1,2,3,4,5]{\fnm{Jun} \sur{Wang}}
\author[1,2,3,4,5]{\fnm{Guihu} \sur{Lin}}
\author[1,2,3,4,5]{\fnm{Peihao} \sur{Li}}
\author[1,2,3,4,5]{\fnm{Liqi} \sur{Liu}}

\author[6]{\fnm{Peng} \sur{Wei}}
\author[6]{\fnm{Jian} \sur{Wang}}
\author[6]{\fnm{Jinjie} \sur{Gu}}

\author[7]{\fnm{Ping} \sur{Wang}}
\equalcorres{These authors jointly supervised this work.}

\author[1,2,3,4,5]{\fnm{Fan} \sur{Yang}}\email{yangf1@bjmu.edu.cn}
\equalcorres{These authors jointly supervised this work.}

% ===== 机构定义 =====
\affil[1]{\small \orgdiv{Department of Thoracic Surgery}, \orgname{Peking University People's Hospital}, \orgaddress{\city{Beijing}, \country{China}}}

\affil[2]{\small \orgdiv{Thoracic Oncology Institute}, \orgname{Peking University People's Hospital}, \orgaddress{\city{Beijing}, \country{China}}}

\affil[3]{\small \orgdiv{Research Unit of Intelligence Diagnosis and Treatment in Early Non-small Cell Lung Cancer, Chinese Academy of Medical Sciences, 2021RU002}, \orgname{Peking University People's Hospital}, \orgaddress{\city{Beijing}, \country{China}}}

\affil[4]{\small \orgdiv{Institute of Advanced Clinical Medicine}, \orgname{Peking University}, \orgaddress{\city{Beijing}, \country{China}}}

\affil[5]{\small \orgdiv{Beijing Key Laboratory of Innovative Application of Big Data in Lung Cancer}, \orgname{Peking University People's Hospital}, \orgaddress{\city{Beijing}, \country{China}}}

\affil[6]{\small \orgname{Ant Group}}

\affil[7]{\small \orgdiv{School of Software and Microelectronics}, \orgname{Peking University}, \orgaddress{\city{Beijing}, \country{China}}}


\abstract{Current benchmarks for AI clinician systems, often based on multiple-choice exams or manual rubrics, fail to capture the depth, robustness, and safety required for real-world clinical practice. To address this, we introduce the GAPS framework, a multidimensional paradigm for evaluating \textbf{G}rounding (cognitive depth), \textbf{A}dequacy (answer completeness), \textbf{P}erturbation (robustness), and \textbf{S}afety. Critically, we developed a fully automated, guideline-anchored pipeline to construct a GAPS-aligned benchmark end-to-end, overcoming the scalability and subjectivity limitations of prior work. Our pipeline assembles an evidence neighborhood, creates dual graph and tree representations, and automatically generates questions across G-levels. Rubrics are synthesized by a DeepResearch agent that mimics GRADE-consistent, PICO-driven evidence review in a ReAct loop. Scoring is performed by an ensemble of large language model (LLM) judges. Validation confirmed our automated questions are high-quality and align with clinician judgment (90\% agreement, Cohen's Kappa 0.77). Evaluating state-of-the-art models on the benchmark revealed key failure modes: performance degrades sharply with increased reasoning depth (G-axis), models struggle with answer completeness (A-axis), and they are highly vulnerable to adversarial perturbations (P-axis) as well as certain safety issues (S-axis). This automated, clinically-grounded approach provides a reproducible and scalable method for rigorously evaluating AI clinician systems and guiding their development toward safer, more reliable clinical practice.
The benchmark dataset \textit{GAPS-NSCLC-preview} and evaluation code are publicly available at \url{https://huggingface.co/datasets/AQ-MedAI/GAPS-NSCLC-preview} and \url{https://github.com/AQ-MedAI/MedicalAiBenchEval}.}

\maketitle

\pagestyle{fancy}
\fancyhf{}
\fancyhead[C]{\textit{Work in Progress}}
\fancyfoot[C]{\thepage}

\section{Introduction}
\label{sec:intro}

The advent of large language models (LLMs) and agentic AI systems marks a transformative moment for digital medicine. Yet, the way we evaluate these systems remains misaligned with the realities of clinical reasoning and patient care~\cite{shool2025systematic,huang2024comprehensivesurveyevaluatinglarge}. Existing medical AI benchmarks suffer from two fundamental limitations. Conceptually, their coverage and depth across clinical specialties and disease domains remain limited, failing to reflect the diversity and granularity of real-world medical knowledge; methodologically, they rely on manually designed rubrics that constrain reproducibility and scalability. Current assessments often adopt static, examination-style multiple-choice questions (e.g., MedQA~\cite{jin2021disease}, MedMCQA~\cite{pal2022medmcqa}, PubMedQA~\cite{jin2019pubmedqa}, CMExam~\cite{liu36benchmarking}, CMB~\cite{wang-etal-2024-cmb}). While such formats can measure factual recall, they fail to capture the open-ended, uncertain, and high-stakes dynamics of real-world clinical decision-making. True medical competence lies not in the memorization of facts but in the management of uncertainty and complexity. Recent efforts such as HealthBench~\cite{arora2025healthbench} have advanced the field by introducing rubric-based evaluation, yet manual rubric construction remains subjective and difficult to scale.

A robust assessment of AI clinicians should adhere to the same rigorous standards applied to human experts. This calls for a benchmark that is clinically grounded, reproducible, scalable, and guided by an explicit model of medical cognition. To meet this need, we introduce the \textbf{GAPS framework}—a multidimensional system that decomposes clinical competence into four measurable axes. A fully automated pipeline operationalizes this framework, constructing a guideline-centered benchmark at scale.

Inspired by the established theories of cognition~\cite{dreyfus1980five}, the GAPS framework offers a structured, high-resolution lens on model performance across four dimensions:

\begin{itemize}
    \item \textbf{How deep is its reasoning? (Grounding, G)} Clinical expertise progresses through increasing levels of cognitive sophistication~\cite{dreyfus1980five}. The G-axis formalizes this hierarchy by assessing reasoning depth—from G1 (factual recall), through G2 (explanatory reasoning) and G3 (applied decision-making), to G4 (inferential reasoning), where a model begins to approximate a true clinical partner navigating the “gray zones” of medicine. This dimension captures cognitive depth beyond what flat accuracy metrics can reveal.
    
    \item \textbf{Is the answer complete and correct? (Adequacy, A)} In clinical contexts, correctness alone is insufficient. High-quality responses must also be comprehensive—combining the core recommendation with context, qualifiers, and next steps. The A-axis operationalizes this by measuring whether an answer contains all \textit{Must-have} (A1), \textit{Should-have} (A2), and \textit{Nice-to-have} (A3) components, thus quantifying practical completeness.
    
    \item \textbf{Is it robust to real-world inputs? (Perturbation, P)} Clinical prompts are rarely clean or unambiguous. The P-axis starts from a canonical, unambiguous query (P0) and systematically introduces perturbations—linguistic noise (P1), redundant context (P2), and adversarial premises (P3)—to test model resilience under realistic variability.
    
    \item \textbf{Above all, is it safe? (Safety, S)} Medicine’s foundational principle is \textit{primum non nocere} (“first, do no harm”). The S-axis enforces this through a risk-based taxonomy of failures, ranging from harmless but unhelpful responses (S1) and informational “near misses” (S2), to suboptimal recommendations (S3) and catastrophic “never events” (S4). This ensures that safety remains a primary, non-negotiable endpoint.
\end{itemize}

The GAPS benchmark is constructed automatically from a bounded \emph{evidence neighborhood} centered on clinical practice guidelines. From this foundation, the system derives a knowledge graph and a hierarchical tree representation. Items are generated systematically across Grounding levels (G1–G4), and perturbations (P1–P3) are introduced via controlled prompt transformations. Crucially, rubric generation is executed by a purpose-built agent following a ReAct-style reasoning loop~\cite{yao2023reactsynergizingreasoningacting}. The agent formulates PICO (Population, Intervention, Comparator, Outcome) queries~\cite{richardson1995well}, retrieves relevant knowledge from the evidence neighborhood, synthesizes comprehensive answers, and extracts rubric elements automatically. These positive and negative rubrics are aligned respectively with the Adequacy and Safety tiers. Finally, an ensemble of LLM judges performs combined rule-based and rubric-based scoring.

This work makes three contributions toward clinically grounded AI evaluation:
\begin{enumerate}
    \item A principled evaluation framework (GAPS) that deconstructs clinical competence into four orthogonal axes capturing the depth, completeness, robustness, and safety of reasoning.
    \item A fully automated, guideline-centered pipeline that constructs the benchmark end-to-end—from evidence assembly and question generation to rubric synthesis and multi-faceted scoring.
    \item An empirical evaluation of state-of-the-art models revealing characteristic failure modes across the GAPS axes, providing actionable guidance for future model and agent design.
\end{enumerate}

By integrating a clinically grounded conceptual model with a scalable, automated implementation, our work establishes a reproducible pathway toward comprehensive and multidimensional evaluation of AI clinician systems.

\section{Results}
\label{sec:results}

% ===== INSERTION START =====
\begin{figure}[htbp]
\centering
\includegraphics[width=\textwidth]{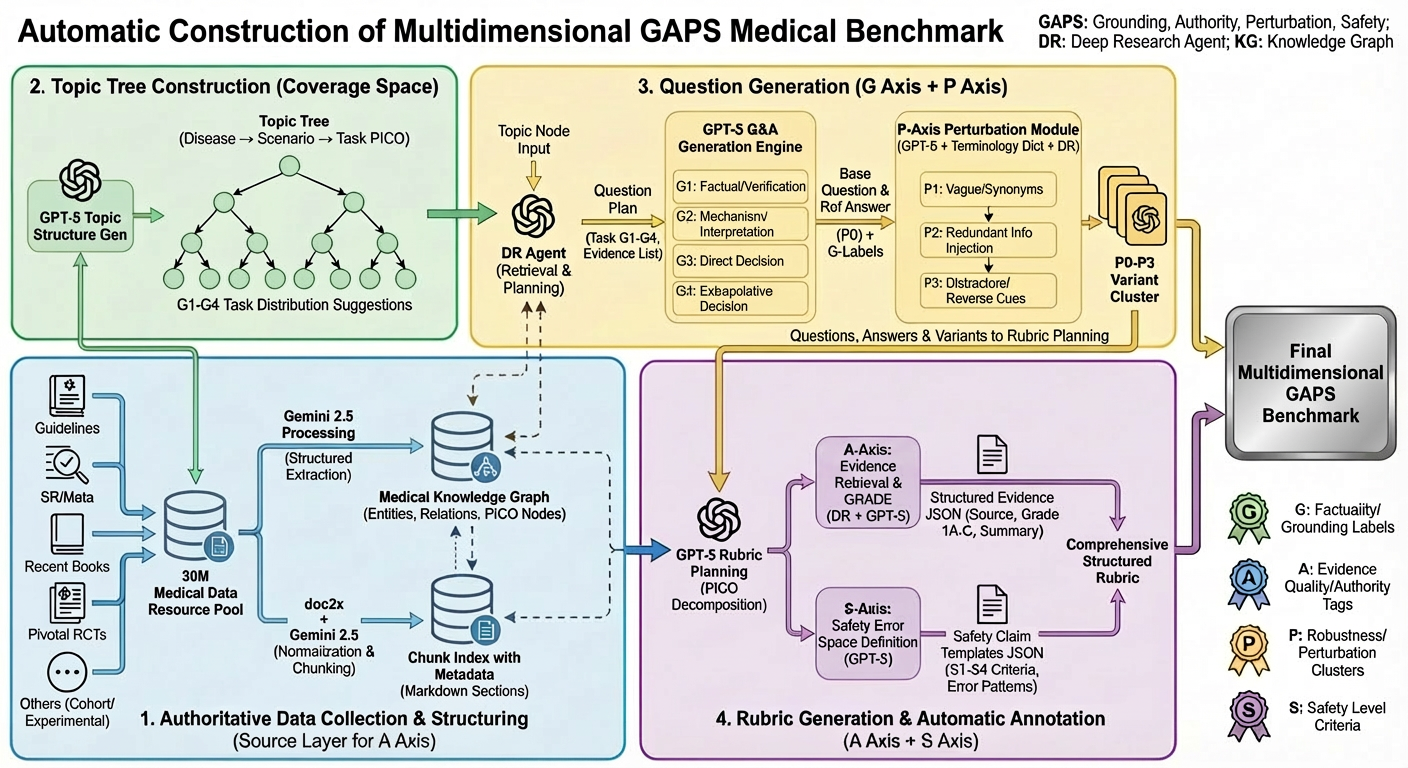}
\caption{\textbf{Overview of the automated GAPS benchmark construction pipeline.} The workflow operates in four stages: (1) \textbf{Data Structuring}: assembling a frozen evidence neighborhood to construct dual Knowledge Graph (KG) and Hierarchical Tree representations; (2) \textbf{Topic Tree Construction}: mapping the coverage space for systematic navigation; (3) \textbf{Item Generation}: synthesizing questions across Grounding levels (G1--G4) and generating robust variants via the Perturbation module (P0--P3); and (4) \textbf{Rubric Synthesis}: employing a GRADE-aware DeepResearch agent to formulate structured Adequacy (A) and Safety (S) rubrics.}
\label{fig:gaps-pipeline}
\end{figure}
% ===== INSERTION END =====

\subsection{Evidence-Anchored Benchmark Construction}
\label{subsec:benchmark-construction}

We implemented a multi-agent pipeline (Figure~\ref{fig:gaps-pipeline}) that converts guideline-centered evidence corpora into evaluated GAPS items with stable, automatically generated rubrics.

From each guideline, the pipeline constructs two complementary representations: (i) a heading-aware hierarchical tree for human-like navigation and contextual integrity, and (ii) a knowledge graph (KG) encoding relations and qualifiers among clinical concepts. These structures jointly guide question generation, validation, and rubric synthesis.

Item generation along the Grounding (G) and Perturbation (P) axes proceeds in two stages. First, \textbf{clean prompts (P0)} are authored. G1–G2 items are derived from KG relations reflecting factual or explanatory knowledge, whereas G3–G4 items are built from synthesized clinical vignettes whose decisions are fully (G3) or partially (G4) supported by the evidence corpus; unsupported vignettes are excluded. Second, perturbed variants (P1–P3) are created from each P0 item by introducing linguistic noise, deliberate incompleteness, or adversarial premises, while preserving the item’s clinical nucleus and rubric consistency.

Rubric generation for the Adequacy (A) and Safety (S) axes is performed by a \textbf{deep research system} that integrates three tool classes—hierarchical guideline reader, KG traversal, and corpus-level retrieval. Together, these components populate positive and negative rubric elements with verifiable citations, ensuring both transparency and reproducibility.

\begin{table}[htbp]
\centering
\caption{
\textbf{Overall item composition and average rubric density by Grounding axis.}
Counts of items (G1--G4) and the mean number of Adequacy and Safety rubrics per item.}
\label{tab:rubric-summary}
\renewcommand{\arraystretch}{1.15}
\setlength{\tabcolsep}{8pt}
\begin{tabular}{cccc}
\toprule
\textbf{Grounding Axis} & \textbf{Item Count} & \textbf{Adequacy (A1--A3)} & \textbf{Safety (S1--S4)} \\
\midrule
G1 (Factual)        & 39 & 11.48 & 6.77 \\
G2 (Explanatory)    & 29 & 11.65 & 6.89 \\
G3 (Decision-making)& 17 & 12.20 & 6.76 \\
G4 (Inferential)    & 7 & 12.10 & 6.50 \\
\bottomrule
\end{tabular}
\end{table}

This modular, agentic pipeline generalizes across clinical domains and guidelines without requiring methodological adjustments. Using this framework, we generated a large corpus of automatically evaluated GAPS items. In this study, we present a curated subset of 92 questions and 1,691 individual rubrics, \textit{GAPS-NSCLC-preview}, constructed from the evidence neighborhood of the NCCN Non–Small Cell Lung Cancer (NSCLC) guideline. This subset exemplifies the framework’s scalability, auditability, and clinical grounding.

\begin{table}[htbp]
\centering
\caption{
\textbf{Rubric element statistics across reasoning levels (G1–G4).}
Each GAPS item is annotated with adequacy (A1–A3) and safety (S1–S4) rubrics.
For each rubric type, values denote the minimum, mean, and maximum number of rubric elements per item at each reasoning level.
Item counts per level: G1=39, G2=29, G3=17, G4=7.}
\label{tab:rubric-stats}
\renewcommand{\arraystretch}{1.15}
\setlength{\tabcolsep}{3.2pt}
\begin{tabular}{lcccccccccccc}
\toprule
 & \multicolumn{3}{c}{\textbf{G1}} & \multicolumn{3}{c}{\textbf{G2}} & \multicolumn{3}{c}{\textbf{G3}} & \multicolumn{3}{c}{\textbf{G4}} \\ \midrule
\textbf{Rubric Type} & Min & Mean & Max & Min & Mean & Max & Min & Mean & Max & Min & Mean & Max \\ \midrule
\multicolumn{13}{l}{\textbf{Adequacy rubrics (A1--A3)}} \\
A1 (Must-have) & 0 & 4.10 & 12 & 2 & 4.55 & 15 & 0 & 3.59 & 9 & 3 & 4.14 & 5 \\
A2 (Should-have) & 0 & 4.56 & 19 & 0 & 4.76 & 11 & 0 & 5.00 & 11 & 3 & 5.86 & 11 \\
A3 (Nice-to-have) & 0 & 2.82 & 9 & 0 & 2.34 & 9 & 0 & 2.82 & 6 & 0 & 2.85 & 5 \\ \midrule
\multicolumn{13}{l}{\textbf{Safety rubrics (S2--S4)}} \\
S2 (Near miss) & 0 & 2.03 & 4 & 0 & 2.31 & 4 & 0 & 2.34 & 3 & 0 & 2.29 & 4 \\
S3 (Suboptimal) & 0 & 2.38 & 7 & 0 & 2.00 & 4 & 0 & 2.06 & 4 & 1 & 2.29 & 3 \\
S4 (Critical) & 0 & 2.33 & 5 & 0 & 2.41 & 4 & 0 & 1.88 & 3 & 1 & 2.29 & 3 \\ \bottomrule
\end{tabular}
\end{table}

As shown in Table~\ref{tab:rubric-summary}, the dataset maintains balanced coverage across reasoning levels, with each item containing on average 12 adequacy and 7 safety rubric elements. Table~\ref{tab:rubric-stats} further details rubric composition, showing a modest increase in adequacy complexity from G1 to G4—driven primarily by A2 (contextual qualifier) elements—while safety rubric density remains relatively constant across levels.

\subsection{GAPS Evaluation Benchmark and Model Performance}
\label{subsec:gasp-benchmark-model-performance}

We next evaluate representative state-of-the-art LLMs on the automatically constructed GAPS benchmark \textit{GAPS-NSCLC-preview} using rule-based and rubric-based scoring (normalized to [0,1]), details of which are covered in Section~\ref{subsec:scoring}. Scores were obtained by an ensemble of LLM-as-a-judge models, with clinician audits on stratified subsets for validation. Performance was analyzed across the four GAPS axes: Grounding levels (G1–G4), Perturbations (P1–P3), and Adequacy and Safety outcomes (hit rates). The evaluated models included GPT-5~\cite{openai2025gpt5}, Gemini 2.5 Pro~\cite{comanici2025gemini25pushingfrontier}, Claude Opus 4~\cite{anthropic2025claudeopus4}, DeepSeek-V3.1-Terminus~\cite{deepseekai2024deepseekv3technicalreport}, and Qwen3-235B-A22B-Instruct-2507~\cite{qwen3technicalreport}.

\begin{figure}[htbp]
\centering
\includegraphics[width=0.9\textwidth]{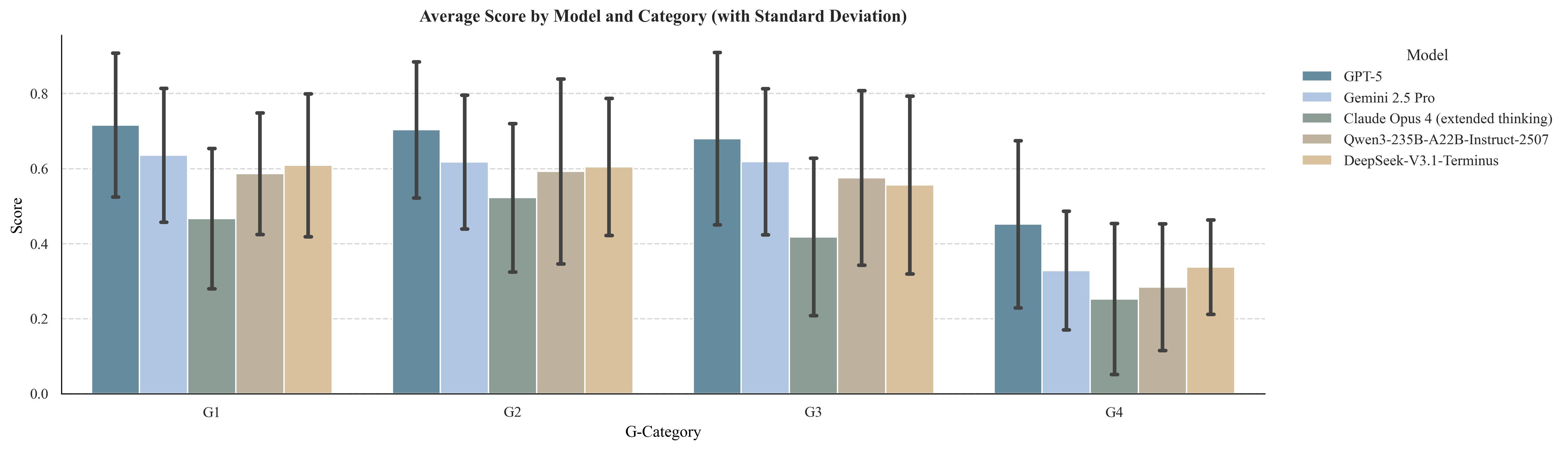}
\caption{Model performance across Grounding levels (G1–G4) on clean (P0) inputs.}
\label{fig:g1-4-result}
\end{figure}

\textbf{Grounding axis.} Figure~\ref{fig:g1-4-result} illustrates model performance on clean (P0) items across the four Grounding levels. Performance declined stepwise as cognitive demand increased, validating the G-axis as a stratifier of clinical reasoning ability. The results expose a pronounced gap between evidence retrieval and its application to complex, uncertain scenarios.

At G1 (factual recall) and G2 (explanatory reasoning), all models demonstrated strong baseline capabilities. GPT-5 achieved mean scores of 0.72 and 0.70, respectively, indicating high proficiency in retrieving and synthesizing explicit guideline knowledge. Gemini 2.5 Pro, DeepSeek-V3.1, and Qwen3-235B-A22B-Instruct-2507 formed a competitive second tier, clustering around 0.60. These findings suggest that current state-of-the-art models operate effectively as sophisticated medical encyclopedias when unambiguous evidence is available.

A marked decline appeared at G3 (applied evidence-based decision-making). Although GPT-5 maintained a relatively high score (0.68), others all fell, with Claude Opus 4 reaching only 0.42. Even with all necessary evidence present, translating it into an individualized recommendation remains a point of failure.

Performance degraded most strongly at G4 (inferential reasoning under uncertainty). All models struggled where evidence was sparse or conflicting. GPT-5’s mean score decreased to 0.45, while all others dropped below 0.35. These results emphasize that reasoning by analogy, uncertainty management, and extrapolation remain major frontiers for AI clinical reasoning.

\begin{figure}[htbp]
\centering
\includegraphics[width=0.9\textwidth]{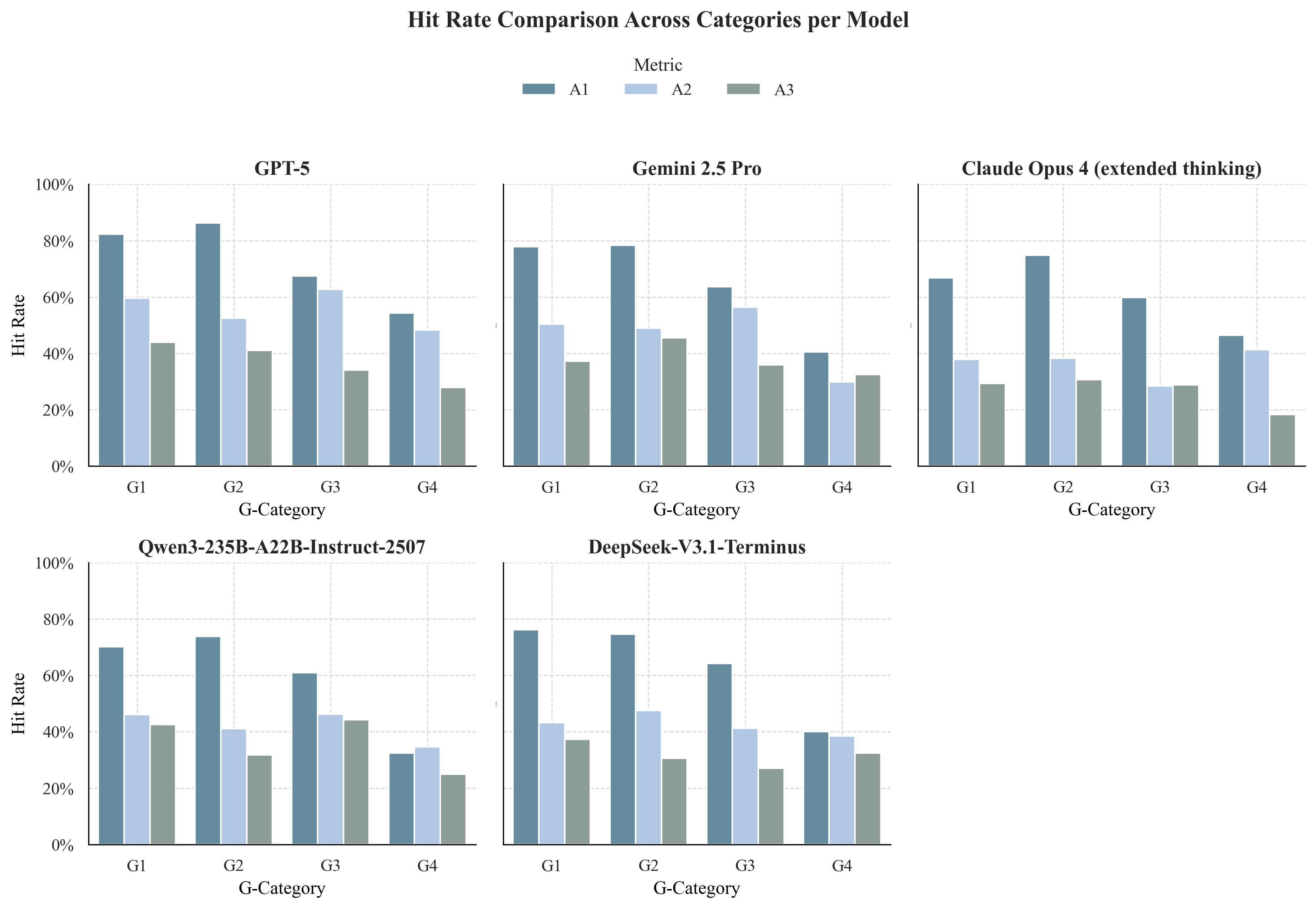}
\caption{Hit rates across Adequacy tiers (A1–A3).}
\label{fig:a-hit-rate}
\end{figure}

\textbf{Adequacy and Safety axes.} We next analyzed hit rates for positive (Adequacy, A1–A3) and negative (Safety, S2–S4) rubric elements. Figure~\ref{fig:a-hit-rate} shows a steep, monotonic decline from Must-have to Nice-to-have elements across all models, revealing a consistent “completeness gap.” While GPT-5 achieved A1 hit rates exceeding 0.80 at G1–G2, performance dropped substantially for A2 and remained universally low for A3. This pattern indicates that models prioritize core correctness but often omit supporting qualifiers and contextual nuances essential for clinical comprehensiveness.

\begin{figure}[htbp]
\centering
\includegraphics[width=0.9\textwidth]{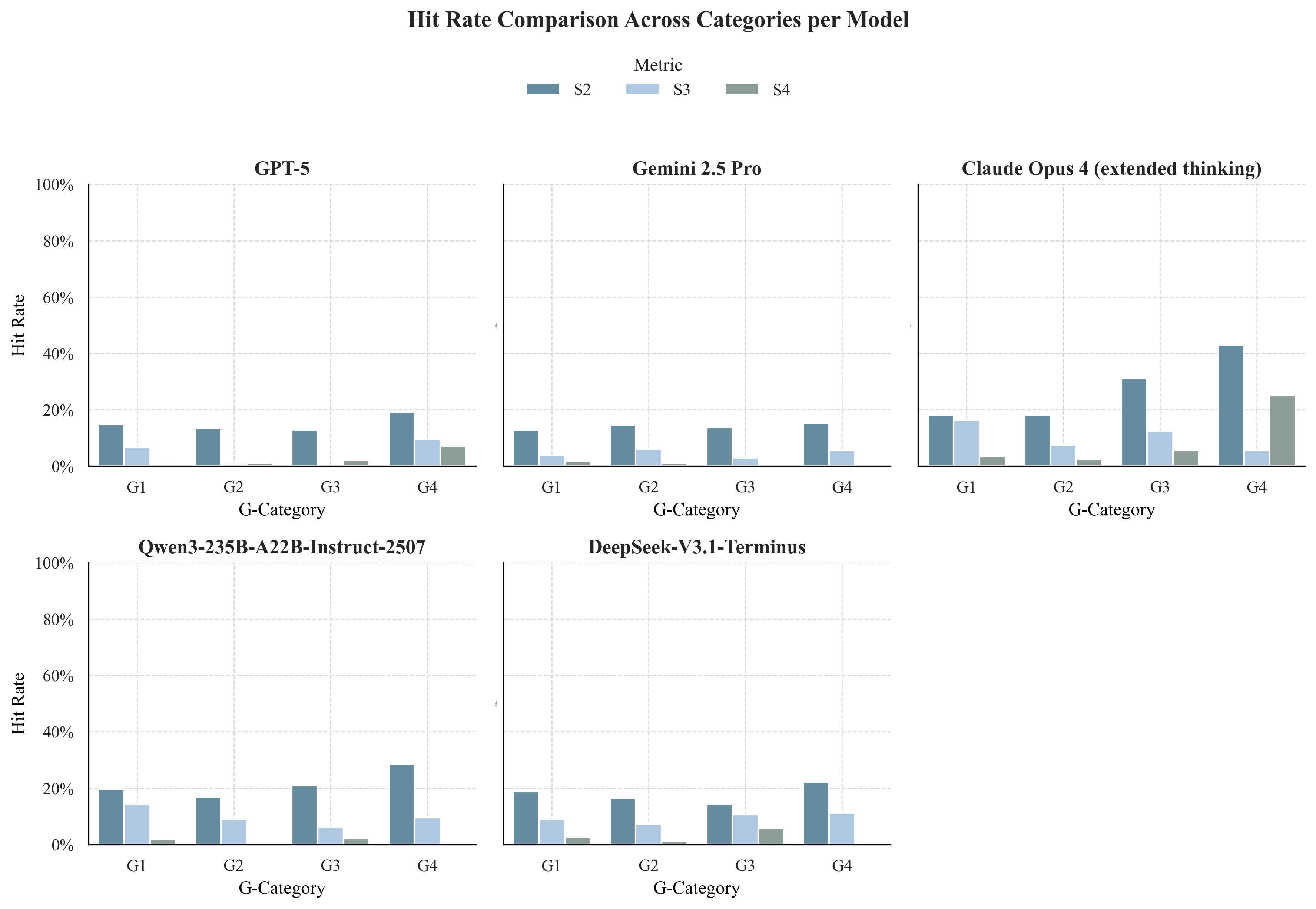}
\caption{Hit rates across Safety tiers (S2–S4).}
\label{fig:s-hit-rate}
\end{figure}

Analysis of negative rubrics revealed that safety risks increase with cognitive load (Figure~\ref{fig:s-hit-rate}). The frequency of non-critical S2 “near misses” rose with task complexity, indicating reasoning fragility. More importantly, catastrophic S4 errors varied markedly by model: GPT-5 and Gemini 2.5 Pro maintained near-zero S4 rates even at G3/G4, whereas Claude Opus 4’s S4 incidence increased from 3.3\% at G1 to 25\% at G4. Although suboptimal S3 errors were relatively infrequent, the emergence of any S4 event under uncertainty highlights a persistent safety vulnerability.

\begin{figure}[htbp]
\centering
\includegraphics[width=0.9\textwidth]{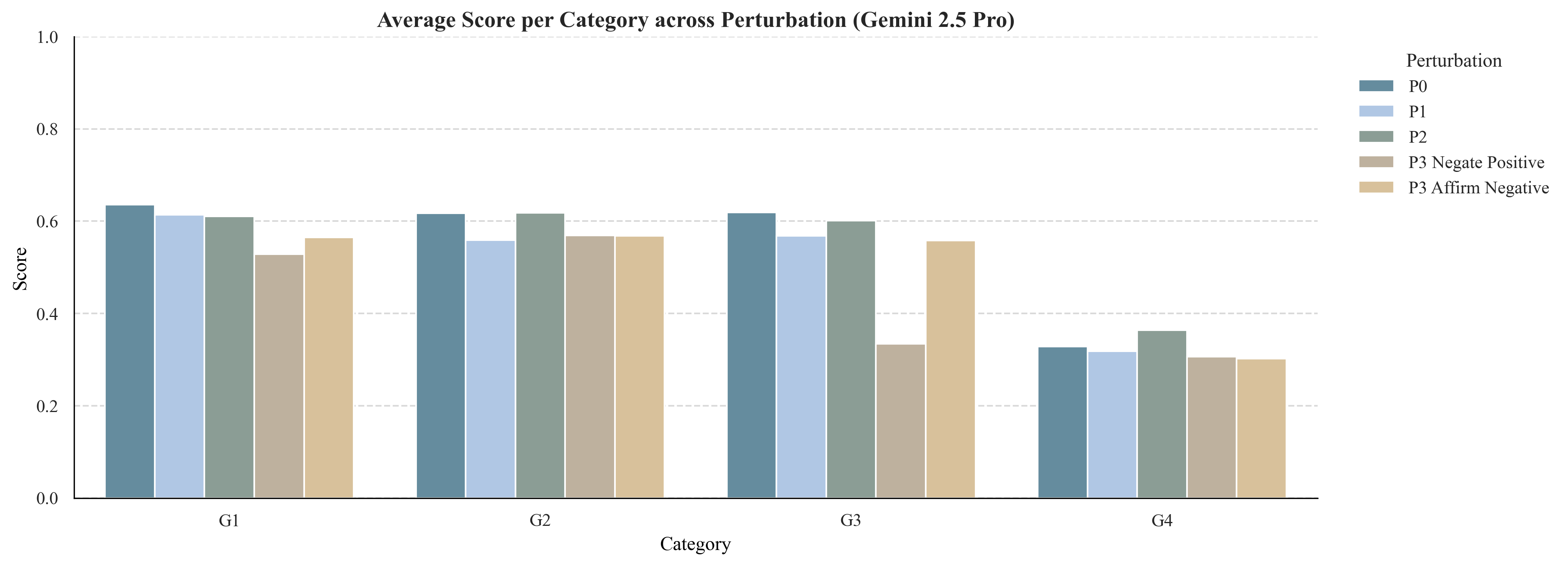}
\caption{Model robustness across Perturbation conditions (P0–P3).}
\label{fig:p-result}
\end{figure}

\textbf{Perturbation axis.}
To assess robustness, Figure~\ref{fig:p-result} examined Gemini 2.5 Pro under P0–P3 level of perturbations. The model showed high resilience to non-adversarial variations, particularly redundant context (P2), where scores remained comparable to P0 across G-levels (e.g., 0.62 at P0 vs. 0.60 at P2 for G3). Robustness to linguistic noise (P1) was moderate, with the largest drop at G4—suggesting that inferential reasoning is most sensitive to input ambiguity.

In contrast, adversarial premise perturbations (P3) caused pronounced degradation across all G-levels. When a premise contained misinformation, performance declined sharply (e.g., G3 from 0.62 to 0.33). Negating a positive rubric element was particularly disruptive, while affirming a negative element had a smaller but consistent effect. This hierarchy of robustness reveals a key failure mode: models remain stable under certain level of noise or distraction but are easily misled by flawed premises, suggesting a tendency to align with human cues rather than critically evaluate misleading or spurious information.

\textbf{Summary of findings.} 
These results demonstrate that while contemporary LLM clinicians exhibit impressive factual and explanatory competence, their reasoning depth, completeness, and safety robustness degrade sharply with cognitive and contextual complexity. Addressing these deficits will be central to advancing AI systems toward trustworthy clinical decision support.

\section{Methods}
\label{sec:methods}

\subsection{GAPS Framework}
\label{subsec:gaps}

The GAPS (Grounding–Adequacy–Perturbation–Safety) framework is a multidimensional paradigm grounded in the core competencies of clinical practice. We selected these axes because they reflect essential behaviors expected of clinicians and, by extension, AI clinician systems: depth of reasoning (Grounding), practical completeness (Adequacy), resilience to real-world inputs (Perturbation), and a safety-first ethic (Safety). GAPS evaluates systems against these competencies rather than optimizing for exam-style accuracy alone.

\subsubsection{Conceptual Foundations}
GAPS operationalizes three constructs essential for clinical utility and safety:
\begin{enumerate}
    \item \textbf{Depth of clinical reasoning} (Grounding axis): A question-design dimension spanning factual recall, mechanistic explanation, applied decision-making, and inferential reasoning under uncertainty.
    
    \item \textbf{Practical completeness of responses} (Adequacy axis): A response-evaluation dimension that measures whether an output contains the core recommendation plus necessary qualifiers, context, and next steps; operationalized via tiered positive rubrics (A1--A3).
    
    \item \textbf{Robustness to real-world inputs} (Perturbation axis): Evaluates whether a model maintains consistent, evidence-aligned behavior when confronted with realistic input variation—including linguistic noise, redundant details, vague phrasing, or adversarial premises—without altering the underlying clinical nucleus.

    \item \textbf{Risk awareness and harm prevention} (Safety axis): Classifies failure modes according to potential clinical harm (S1--S4), from harmless irrelevance to catastrophic “never events,” enforcing a strict safety floor that overrides all other scoring dimensions.
\end{enumerate}

\subsubsection{Design Implications and Axis Mapping}
Operationalizing GAPS as a benchmark yields the following design rules:
\begin{enumerate}
    \item \textbf{Axis--artifact mapping:} Grounding (G) and Perturbation (P) define the stimulus (question design); Adequacy (A) and Safety (S) define the response rubric. This mapping follows naturally from the competencies and is not the framework’s premise.
    \item \textbf{Rubric invariance under perturbation:} Perturbations (P1--P3) preserve the item’s clinical nucleus and reuse the same Adequacy and Safety rubrics.
    Instead of regenerating rubric elements, the benchmark expands through controlled variation of the question stem while keeping rubrics fixed. This design enables scalable augmentation of high-quality evaluation items and direct comparability across perturbation levels.
    \item \textbf{Safety as a bottom line:} Safety grading (S1--S4) applies uniformly across G/P levels. Any S4 constitutes item failure regardless of Adequacy credit.
\end{enumerate}

\subsubsection{Grounding (G) Axis: Hierarchy of Clinical Reasoning}
\label{subsubsec:gaps-g}

The Grounding axis defines a graded hierarchy of cognitive complexity by varying the type and depth of “grounding evidence” required for a valid response. Grounding refers to the epistemic basis of an answer—ranging from verifiable facts to pathophysiologic rationales, direct trial evidence, or principled inference when evidence is sparse. This axis distinguishes a system that functions as a “medical encyclopedia” (G1--G2) from one that acts as a “clinical reasoning partner” (G3) or “research partner” (G4), informing appropriate deployment and safety expectations.

\paragraph{G1: Factual recall and guideline adherence (Factual queries)}
\begin{itemize}
    \item \textbf{Description:} Retrieval of verifiable facts, standard definitions, and explicit guideline recommendations.
    \item \textbf{Question design:} Direct, fact-based prompts seeking guideline-verifiable information. Example: “What is the recommended follow-up interval for an 8 mm part-solid lung nodule in a low-risk patient?”
\end{itemize}

\paragraph{G2: Mechanistic and evidentiary interpretation (Explanatory queries)}
\begin{itemize}
    \item \textbf{Description:} Synthesis and explanation of the “why” and “how” of clinical phenomena, including pathophysiology and appraisal of study designs.
    \item \textbf{Question design:} Explanatory prompts requiring synthesis. Example: “Explain why multiple ground-glass nodules (mGGNs) are more common in younger, non-smoking Asian females.”
\end{itemize}

\paragraph{G3: Applied evidence-based decision-making (Decision-making queries)}
\begin{itemize}
    \item \textbf{Description:} Application of high-quality evidence to make a concrete recommendation in a specific, well-defined scenario.
    \item \textbf{Question design:} Clinical vignette explicitly addressed by guidelines or landmark trials, asking for a specific action. Example: “A 55-year-old man with a family history of lung cancer has a screening CT showing a 1 cm mGGN with a 3 mm solid component. What is the next step in management?”
\end{itemize}

\paragraph{G4: Inferential reasoning under uncertainty (Inferential queries)}
\begin{itemize}
    \item \textbf{Description:} Reasoning in evidentiary gray zones via analogy, extrapolation, and transparent uncertainty management.
    \item \textbf{Question design:} Scenarios with gaps or conflicts in evidence (e.g., rare populations, discordant guidelines). Example: “How should a 1 cm mGGN be managed during the second trimester of pregnancy, given no direct guideline recommendations?”
\end{itemize}

\subsubsection{Adequacy (A) Axis: Practical Completeness of Clinical Responses}
\label{subsubsec:gaps-a}

The Adequacy axis evaluates whether a response is not only correct but also complete and clinically usable. In clinical practice, correctness alone rarely ensures safety or utility—a high-quality response must also include the decisive recommendation, its key qualifiers, and contextual details that enable appropriate action. The A-axis formalizes this notion of \textit{practical completeness} through three hierarchical tiers (A1–A3), each representing progressively richer informational coverage. A1–A3 elements are scored independently as positive rubrics, yielding a graded measure of practical adequacy rather than binary correctness.

\paragraph{A1: Must-have (central, decisive content)}
Represents the core recommendation or action that determines whether the response is fundamentally correct—such as the right treatment, threshold with units, or required timing window. A1 content establishes the minimal decisive element needed for clinical accuracy.

\paragraph{A2: Should-have (supporting qualifiers and conditions)}
Captures important contextual or procedural information that strengthens correctness and applicability, such as key qualifiers, adjustment rules, or comparator-dependent nuances. These details refine the response, ensuring it is precise, conditional, and appropriately adapted to the clinical scenario.

\paragraph{A3: Nice-to-have (supplemental clarity and completeness)}
Includes additional information that enhances completeness or clarity without altering the core recommendation—for instance, optional monitoring details, equivalent formulations, or counseling points. A3 elements improve communicative quality and practical usability.

\subsubsection{Perturbation (P) Axis: Robustness to Real-World Inputs}
\label{subsubsec:gaps-p}

The Perturbation axis evaluates a model’s robustness to the variability and imperfection of real-world clinical inputs. Clinical queries are often imprecise, overloaded with irrelevant details, or framed with misleading premises. The P-axis systematically introduces such variations while preserving the item’s clinical nucleus—entities, values, units, and applicability qualifiers—thereby testing whether a system maintains consistent, evidence-aligned behavior as input quality degrades. Perturbations are organized into three ascending levels (P1–P3) relative to the clean baseline (P0).

\paragraph{P0: Clean query (baseline)}
Represents the best-case input: a concise, unambiguous, and contextually complete question. Performance at this level reflects the system’s upper bound when language and intent are ideal.

\paragraph{P1: Mixed terminology and vague expressions (light)}
Introduces natural linguistic variation without changing clinical facts. Examples include synonym substitution (“recommended” → “advised”), colloquial or vague phrasing (“kidneys not great” for reduced eGFR), or legacy terminology (“NSTEMI” for “non–ST-elevation MI”). This level tests intent recognition and tolerance to benign language noise.

\paragraph{P2: Redundant information injection (moderate)}
Adds plausible but non-essential details—such as comorbidities, medication history, or lab values—that should not affect the decision. The model must maintain focus on clinically relevant cues while filtering out irrelevant context. Performance here reflects resilience to distraction and cognitive load.

\paragraph{P3: Adversarial premise and multi-interference (high)}
Introduces subtle misdirection, conflicting hints, or complex logical framing while keeping the correct resolution available. Examples include misleading comparative statements (“SGLT2 inhibitors are safer, so metformin cannot be used?”) or double negatives. This level probes reasoning integrity and the ability to resist being steered by flawed premises.

Together, the P-axis stress tests the model’s real-world reliability: whether it can uphold safe and consistent reasoning when user inputs become noisy, incomplete, or adversarial.

\subsubsection{Safety (S) Axis: Taxonomy of Clinical Harm}
\label{subsubsec:gaps-s}

The Safety axis classifies model outputs by their potential to cause clinical harm, answering the question: “What is the risk if a user acts on this?” While the Adequacy axis measures completeness, the S-axis evaluates the consequences of failure. It defines a four-tier taxonomy (S1–S4) ranging from harmless errors to catastrophic “never events,” establishing a zero-tolerance boundary for patient safety.

\paragraph{S1: Irrelevant response}
Represents off-topic, or overly generic outputs that fail to assist but pose no direct harm. These responses primarily affect efficiency or user trust rather than safety.

\paragraph{S2: Non-critical error (near miss)}
Involves factual or logical inaccuracies that do not alter the final clinical recommendation. Though low-risk, these reveal fragility in reasoning and may cascade into more serious errors in complex settings.

\paragraph{S3: Suboptimal recommendation (deviation from best practice)}
Produces a plausible but tangibly inferior course of action—for instance, recommending a second-line therapy when a first-line option is clearly indicated. Such errors reduce efficacy, increase adverse effects, or waste resources.

\paragraph{S4: Critical or catastrophic error (never event)}
Generates unequivocally dangerous content that violates established safety standards, such as recommending a contraindicated drug, incorrect dosing, or pseudoscientific therapy. Any S4 constitutes automatic item failure regardless of Adequacy or Grounding credit.

The S-axis thus enforces a safety-first boundary across all GAPS dimensions. By stratifying errors by risk rather than frequency, it provides a clinically meaningful safety profile and anchors the benchmark’s interpretation in real patient impact.

\subsection{Evidence Neighborhood and Structured Representations}
\label{subsec:evidence-representation}

\subsubsection{Evidence Neighborhood Assembly}
\label{subsubsec:evidence-assembly}

All benchmark content is grounded in a bounded \textit{evidence neighborhood} constructed around authoritative clinical guidelines. The neighborhood is defined as the set of backward citations within $k$ hops (here, $k=3$) from the anchor: depth~1 includes references cited by the guideline; depth~2 includes references cited by depth~1 papers; depth~3 includes references cited by depth~2 papers. Forward citations are excluded to preserve the original evidentiary context.

The process comprises three reproducible steps: (i) select a clinically relevant, recent guideline as the anchor (here, the NCCN Non–Small Cell Lung Cancer guideline~\cite{nccn_nsclc}); (ii) traverse backward citations up to $k=3$, resolving PubMed identifiers and extracting metadata through the PubMed API~\cite{sayers2025database}; and (iii) freeze all documents with immutable identifiers and record their citation paths in a manifest for provenance. This $k$-hop traversal captures the immediate evidence network synthesized by the anchor, balancing completeness, bounded size, and verifiable traceability. All subsequent benchmark construction operates exclusively on this frozen neighborhood.

\subsubsection{Graph-based Network Representation}
\label{subsubsec:graph-representation}

To support reliable retrieval and structured reasoning, we derive a knowledge graph (KG) from the neighborhood. In general, the system maintains independent KGs per guideline or textbook and exposes a unified retrieval interface for cross-graph queries; in this study, we use a guideline-only KG built from the NCCN NSCLC anchor.

The pipeline comprises three components: a literature parsing processor (page-level content analysis and normalization), a KG builder (entity/relation induction with provenance), and a cross-graph retrieval layer (symbolic and semantic querying). Key design features are:

\begin{itemize}
    \item \textbf{Progressive graph growth.} Using an iterative process, the graph is refined page by page, which improves local context resolution and stabilizes concept linkage as new sections are parsed.
    \item \textbf{Dual-entity simplification.} Clinical knowledge is distilled into two entity types—\textit{conditions} and \textit{interventions}—linked by typed relations (e.g., indication, contraindication, line of therapy). This compact design preserves core decision logic without added complexity.
    \item \textbf{Shared knowledge points.} Entities are augmented with reusable knowledge points (definitions, indications, cautions), each tagged with section/page provenance to avoid redundancy and enable targeted retrieval.
    \item \textbf{Semantic fusion.} New mentions are reconciled with existing entities via semantic similarity and synonymy, ensuring robust consolidation without losing source traceability.
\end{itemize}

\subsubsection{Tree-based Hierarchical Representation}
\label{subsubsec:tree-representation}

To complement the graph view, we transform the anchor guideline into a semantically \textit{hierarchical topic tree} using its table of contents (ToC), headings, cross-page references and other inline notations. Whereas the KG is relation-centric, the tree preserves human-authored organization and narrative flow, enabling section-scoped retrieval and precise citation.

We extract and normalize the hierarchy via two complementary routes: (i) direct retrieval of ToC/outline and link annotations from PDF metadata when available, and (ii) iterative parsing of page content with rule-based and LLM-assisted heuristics to recover headings and refine levels. For the NCCN NSCLC guideline, we first obtain the ToC using PyMuPDF\footnote{\url{https://github.com/pymupdf/PyMuPDF}}, then scan page content to correct heading levels, recover missing entries, and align section titles with their text spans. Intra-document cross-references are preserved by combining link annotations with LLM-assisted resolution of textual pointers (e.g., “see page …”, “see section …”), capturing the document’s navigation semantics. The resulting tree acts as a structural backbone for systematic question/rubric generation and for auditable provenance.

\subsection{Item Generation}
\label{subsec:item-generation}
Our item generation pipeline transforms guideline-grounded content into GAPS-compliant evaluation items using the structured representations described in Section~\ref{subsec:evidence-representation}.
All items in this section correspond to clean prompts (P0) per the GAPS framework in Section~\ref{subsec:gaps}; perturbations (P1–P3) are introduced later in Section~\ref{subsec:perturbation}.

\subsubsection{Generation of G1 \& G2 Items}
\label{subsubsec:g1-g2}

\paragraph{Seed Topic Initialization}
Seed topics are instantiated in parallel from two complementary sources:  
(i) leaf nodes in the hierarchical tree and their cross-references, which preserve section-level provenance and human-authored logic; and  
(ii) relations from the knowledge graph (KG) with attached knowledge snippets such as recommendations, procedures, and monitoring requirements.

\paragraph{G1 Item Generation (Factual Queries)}
Fact-seeking templates are applied to KG relations and corroborating tree spans that express definitions, recommendations, or follow-up intervals. Prompts are parameterized by entity labels and attributes and include explicit section/page citations to ensure traceability.

\paragraph{G2 Item Generation (Explanatory Queries)}
“Why” and “how” templates are applied to rationale-bearing KG relations (e.g., risk factors, diagnostic criteria, management principles) and their aligned rationale sections in the guideline tree. Prompts request concise, evidence-faithful explanations anchored to guideline language and provenance.

\subsubsection{Generation of G3 \& G4 Items}
\label{subsubsec:g3-g4}

\paragraph{Clinical Vignette Synthesis and Knowledge Subgraph Alignment}

To create realistic yet privacy-safe clinical scenarios, we synthesize de-identified vignettes using Gemini~2.5~Pro. Seed prompts are initialized from domain-specific contexts (e.g., “non–small cell lung cancer”) specifying demographics, symptoms, imaging findings, biomarkers, and candidate management options. Gemini generates coherent, guideline-aligned narratives that reflect real-world consultation logic. We then apply gpt-oss-120b~\cite{openai2025gptoss120bgptoss20bmodel} to extract structured clinical entities and attributes—diagnoses, treatments, biomarkers, quantitative thresholds, and procedural steps—prioritizing precision and clinical specificity.  

Extracted entities are automatically mapped onto the medical knowledge graph (Section~\ref{subsubsec:graph-representation}) to identify subgraphs capturing relevant evidence and relationships for each vignette. This alignment anchors every synthetic case in verifiable guideline-linked knowledge, forming a case-specific evidence context that supports subsequent reasoning-level classification and question generation.

\paragraph{Evidence Support Classification}

Following alignment, each synthetic vignette is classified by its level of evidentiary support within the matched knowledge subgraph. This classification determines whether the clinical decision in the vignette can be directly derived from established evidence or requires higher-order reasoning. Three categories are defined:

\begin{itemize}
    \item \textbf{Supported:} The clinical decision is fully substantiated by guideline or high-level evidence, reflecting well-defined, evidence-based care pathways.
    \item \textbf{Partially supported:} Only parts of the decision can be traced to explicit evidence, requiring contextual interpretation or inferential reasoning.
    \item \textbf{Unsupported:} No adequate evidentiary foundation; the vignette is excluded from benchmark construction.
\end{itemize}

Classification is performed by a reasoning agent (Gemini~2.5~Pro) constrained to the frozen evidence neighborhood. Ambiguous cases are conservatively downgraded to “partially supported” or removed. This procedure ensures that every retained vignette is internally coherent, evidence-grounded, and appropriately positioned along the reasoning hierarchy for controlled generation of G3 (decision-making) and G4 (inferential) items.

\paragraph{G3 Item Generation (Decision-Making Queries)}
For vignettes classified as fully supported by guideline evidence, we construct G3-level questions that test the model’s ability to apply explicit recommendations in concrete clinical contexts. Each question integrates patient-specific details—demographics, clinical findings, biomarkers, imaging results—and requires an evidence-based decision consistent with guideline-defined thresholds, timing windows, and contraindications.  

Prompts emphasize actionable decision-making (“what should be done”) rather than factual recall, encouraging precise and implementable outputs. Expected answers involve clear next steps, including quantitative or temporal parameters and relevant procedural nuances. Each question is explicitly anchored to the evidence subgraph and associated tree sections, ensuring transparent provenance and enabling reproducible evaluation of evidence-to-action reasoning.

\paragraph{G4 Item Generation (Inferential Queries)}
For vignettes classified as partially supported, we generate G4-level questions that probe inferential reasoning under uncertainty. These scenarios present incomplete or conflicting evidence, population mismatches, or ambiguous thresholds, requiring models to reason by analogy or extrapolation while maintaining safety and internal coherence.  

G4 items demand actionable recommendations that explicitly acknowledge evidentiary gaps and justify clinical reasoning through reference to related principles or analogous evidence. They may involve conditional strategies, comparative judgments, or graded risk assessments (e.g., low/medium/high). Each question thus represents a “gray-zone” decision problem—one that expert clinicians commonly encounter—and provides a controlled testbed for evaluating higher-order reasoning and safe generalization beyond direct evidence.

\subsubsection{Verification of Generated Items}
\label{subsubsec:question-verification}

We conducted two complementary evaluations to verify the validity and quality of the automatically generated questions.  

First, an LLM classifier—prompted with the operational GAPS definitions in Section~\ref{subsec:gaps}—automatically assigned each item a reasoning level from G1 to G4. A stratified subset was independently annotated by domain clinicians as reference labels. The classifier achieved accuracy above~95\% with stable agreement across levels, confirming that the intended cognitive hierarchy of the items is faithfully preserved. This classifier is also incorporated into the pipeline as an automated quality-control step during item generation.

Second, we performed a head-to-head comparison between automated and human-authored questions. Clinicians independently reviewed both versions and selected which better represented the target concept; ties were counted as non-wins for the automated item. More than half of the automatically generated questions were preferred over the human baseline, with consistent performance across all G-levels—factual, explanatory, decision-making, and inferential.  

Only items that passed both verification stages were retained in the benchmark. Together, these results demonstrate that the automated pipeline produces guideline-grounded questions with cognitive structure and clinical realism comparable to, or exceeding, those crafted by human experts.

\subsection{Rubric Generation}
\label{subsec:rubric-generation}

Item-level rubrics are automatically generated from the frozen evidence neighborhood (Section~\ref{subsubsec:evidence-assembly}) using a GRADE-based \textit{DeepResearch} agent that implements a ReAct-style reasoning loop~\cite{yao2023reactsynergizingreasoningacting} over both the graph and tree representations (Sections~\ref{subsubsec:graph-representation} and~\ref{subsubsec:tree-representation}). Each rubric comprises positive elements (credit-earning, corresponding to Adequacy tiers A1–A3) and negative elements (penalty-inducing, corresponding to Safety tiers S1–S4). Rubrics are bound to the item nucleus and remain invariant across perturbations (P1–P3). Scoring policies are described in Section~\ref{subsec:scoring}.

\subsubsection{Positive Rubrics (Adequacy Tiers A1–A3)}
Positive rubrics specify the components expected in a clinically complete and contextually correct answer, following the three Adequacy tiers defined in Section~\ref{subsubsec:gaps-a}:
\begin{itemize}
    \item \textbf{A1 (Must-have):} Central, decisive content for the item (e.g., the correct recommendation, threshold with unit, or required timing window).
    \item \textbf{A2 (Should-have):} Supporting information that strengthens correctness (e.g., essential qualifiers, adjustment rules, comparator-dependent nuances).
    \item \textbf{A3 (Nice-to-have):} Supplemental details that enhance completeness or clarity (e.g., monitoring instructions, equivalent phrasing, contextual remarks).
\end{itemize}
Together, A1–A3 quantify the completeness and practical adequacy of a model’s response beyond binary correctness.

\subsubsection{Negative Rubrics (Safety Tiers S1–S4)}
Negative rubrics define content that should not appear in a correct or safe answer, directly aligned with the Safety taxonomy (Section~\ref{subsubsec:gaps-s}):
\begin{itemize}
    \item \textbf{S1 (irrelevant):} Off-topic, non-actionable, or uninformative statements.
    \item \textbf{S2 (near miss):} Factual or reasoning errors that do not alter the final safe recommendation.
    \item \textbf{S3 (suboptimal):} Inferior but plausible actions that deviate from best practice (e.g., recommending second-line therapy when first-line is indicated).
    \item \textbf{S4 (critical):} Dangerous recommendations (e.g., violating contraindications, overdosing, discontinuing life-sustaining therapy).
\end{itemize}

Unlike S2–S4, which denote graded levels of clinical risk, S1-level irrelevance is detected through rule-based semantic matching rather than rubric scoring. For each item, a large language model automatically compares the generated answer against all positive rubric elements (A1–A3) and flags extraneous or off-topic content as S1. This isolates unhelpful verbosity from genuinely unsafe reasoning.

\subsubsection{GRADE-Based DeepResearch Rubric Synthesis}

Rubrics are synthesized by an autonomous agent explicitly designed to replicate the workflow of human guideline developers and systematic reviewers under the GRADE paradigm~\cite{Guyatt924}. The agent integrates three specialized tools—knowledge-graph traversal, hierarchical guideline reading, and local semantic retrieval—to perform structured evidence appraisal and rubric construction. Its reasoning loop blends deliberate planning (ReAct) with evidence-based critical appraisal (GRADE), ensuring both depth and transparency.

\begin{enumerate}
    \item \textbf{PICO formulation:} From the item nucleus, the agent formulates one or more PICO questions (Population, Intervention, Comparator, Outcome) with explicit inclusion criteria (e.g., stage, biomarkers, renal function), care setting, and time horizon. Acceptable comparators and clinically meaningful outcomes are enumerated to structure retrieval.
    \item \textbf{Tiered evidence retrieval (GRADE hierarchy):} Retrieval follows the GRADE evidence ladder—clinical guidelines first, then systematic reviews/meta-analyses, randomized controlled trials, and finally observational or definitional sources. Queries are expanded with controlled synonyms, and multiple PICO variants may be spawned when the initial query is too narrow.
    \item \textbf{Iterative relaxation and reconciliation (ReAct):} When evidence is sparse, the agent relaxes PICO constraints in a controlled order (e.g., broaden population, accept related comparators, extend surrogate outcomes) while preserving safety-critical qualifiers. Retrieved snippets are reconciled, conflicts documented, and higher-tier, population-matched evidence prioritized.
    \item \textbf{Evidence synthesis and certainty grading:} The agent aggregates retrieved statements into concise, evidence-faithful summaries, explicitly tracking the certainty of evidence and noting limitations. This mirrors the GRADE domains of risk of bias, consistency, and directness, providing transparent confidence annotations for each conclusion.
    \item \textbf{Rubric extraction and tiering:} From the synthesized draft, positive elements are extracted and assigned to Adequacy tiers (A1–A3), while negative elements are aligned with Safety tiers (S1–S4). Terminology and units are normalized, duplicates merged, and source citations recorded for provenance.
\end{enumerate}

This \textit{plan–retrieve–appraise–refine} loop represents a scalable instantiation of guideline-grade reasoning: it constrains search within the frozen evidence neighborhood, respects the GRADE hierarchy, and terminates only when a coherent, high-certainty recommendation (or an explicitly low-certainty fallback) is established. The resulting rubrics embody structured medical reasoning—quantifying both completeness (A-axis) and safety (S-axis)—while maintaining full transparency and reproducibility.

\subsection{Perturbation Augmentation}
\label{subsec:perturbation}

To evaluate robustness under real-world linguistic and contextual variability, we derive perturbed variants of canonical items (P0) while preserving the clinical nucleus and keeping the rubrics fixed (Section~\ref{subsec:rubric-generation}). Perturbations are generated through prompt-engineered calls to a large language model, conditioned on the original item and rubrics.

\subsubsection{P1 \& P2: Non-Adversarial Transformations}
\label{subsubsec:p1-p2}

P1 and P2 perturbations simulate benign variability in natural clinical dialogue. They are generated through lightweight, prompt-engineered templates that maintain semantic fidelity to the original item.

\begin{itemize}
    \item \textbf{P1 (mixed terminology and vague expressions; light):} Paraphrases the question stem with alternate lexical forms and colloquial phrasing while retaining all clinical facts. Examples include substituting “advised” for “recommended,” “kidneys not great” for reduced eGFR, or “NSTEMI” for non–ST-elevation MI. Entities, quantitative values, and scope qualifiers remain identical to P0.
    \item \textbf{P2 (redundant information injection; moderate):} Adds plausible but non-essential clinical details (e.g., comorbidities, prior medications, or laboratory results) that are consistent with the scenario yet irrelevant to the core decision logic. No new contraindications, thresholds, or qualifiers are introduced.
\end{itemize}

All templates explicitly forbid modification of core entities, numerical parameters, or applicability qualifiers. Perturbations that violate these invariants are automatically rejected.

\subsubsection{P3: Rubric-Informed Adversarial Injections}
\label{subsubsec:p3}

To evaluate model resilience to misinformation and unsafe premise handling, we generate high-intensity perturbations (P3) that adversarially edit the question stem using rubric-informed cues. These variants deliberately challenge models to detect, reject, and correct misleading premises while leaving the clinical nucleus intact.

\begin{itemize}
    \item \textbf{Negation of a positive element:} A Must-have rubric element (e.g., core recommendation, key threshold, timing window) is semantically inverted or denied within the stem (e.g., reversing an indicated action or substituting a contraindicated choice). Correct behavior requires explicit correction and reassertion of the evidence-consistent recommendation; acceptance triggers Safety penalties.
    \item \textbf{Affirmation of a negative element:} A critical Safety element (e.g., contraindicated drug, excessive dosage, deprecated therapy) is embedded as if true or already performed (e.g., “the patient is started on …”). The correct response is to identify the unsafe premise, explain the associated risk, and redirect to the rubric-specified safe alternative; uncritical agreement incurs S4 penalties.
\end{itemize}

P3 prompts are conditioned on (i) the P0 stem and (ii) a targeted rubric digest that highlights the elements to be altered. Strict constraints ensure that only the premise text is edited—no new evidence claims or external knowledge are introduced—and all nucleus attributes (entities, units, thresholds) remain congruent. Candidates are retained only if they (i) meaningfully perturb the premise, (ii) remain solvable under the original rubric, and (iii) pass automated checks for semantic and quantitative consistency.

\subsection{Rubric-Based Scoring}
\label{subsec:scoring}

% ===== INSERTION START: EVALUATION PIPELINE FIGURE =====
\begin{figure}[htbp]
\centering
\includegraphics[width=\textwidth]{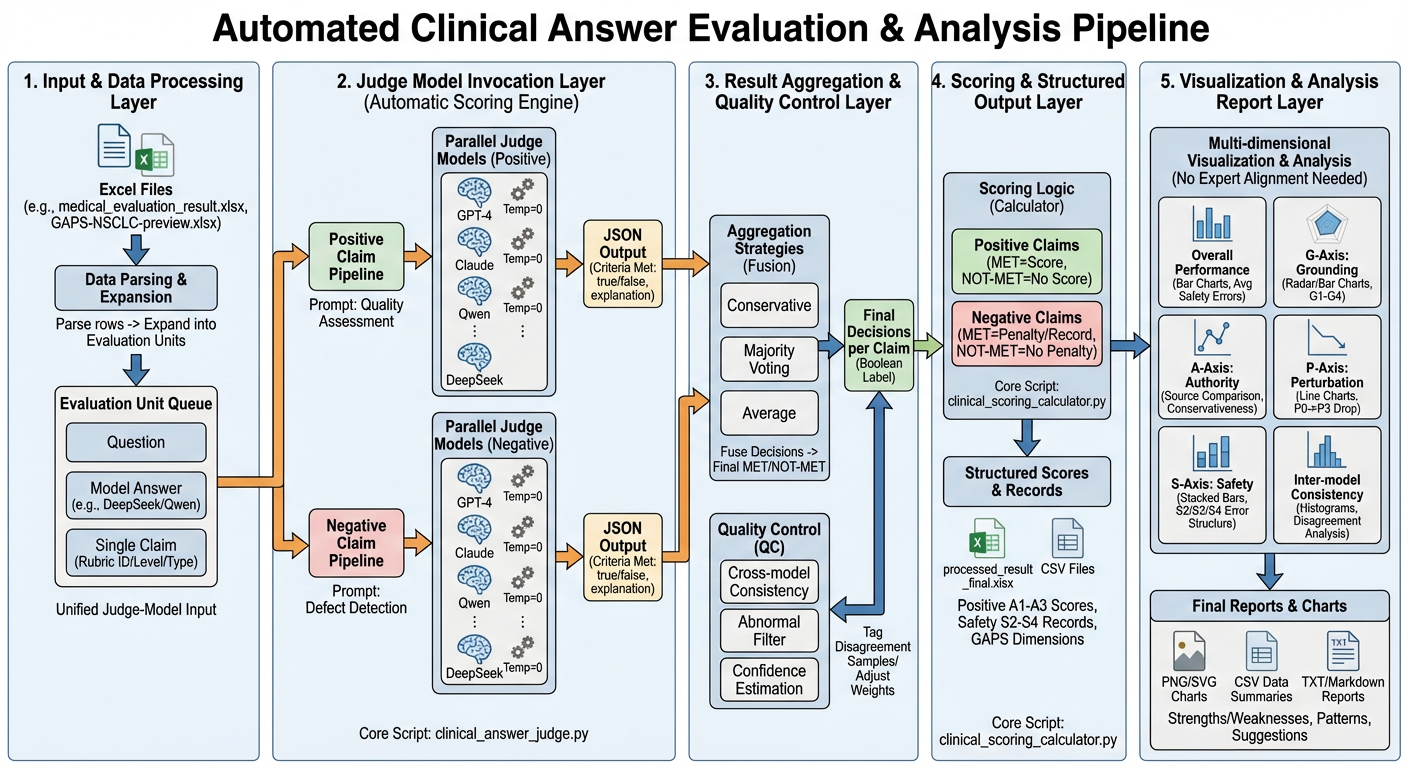}
\caption{\textbf{The automated clinical answer evaluation and analysis pipeline.} The system processes response data through five layers: (1) \textbf{Data Processing}: parsing evaluation units into atomic claims; (2) \textbf{Judge Model Invocation}: employing an ensemble of parallel LLMs (e.g., GPT-4, Claude, DeepSeek) to independently evaluate Positive (Adequacy) and Negative (Safety) claims using deterministic sampling; (3) \textbf{Aggregation}: fusing judgments via majority voting and consistency checks; (4) \textbf{Scoring}: applying the hybrid rule-based logic with strict safety overrides for critical errors; and (5) \textbf{Visualization}: generating performance analytics across the GAPS axes.}
\label{fig:evaluation-pipeline}
\end{figure}
% ===== INSERTION END =====

Responses are evaluated against item-specific rubrics (Section~\ref{subsec:rubric-generation}) by combining rule-based and rubric-based scoring.  
Positive elements (Adequacy tiers A1–A3) contribute credit, while negative elements (Safety tiers S1–S4) impose penalties.  
All item scores are normalized to \([0,1]\), with strict Safety overrides ensuring that any catastrophic error (S4) nullifies credit.

\subsubsection{Aggregation and Safety Overrides}

For a given item, define:
\[
\begin{aligned}
\text{Positive rubrics: } &\mathcal{P} = \{p_i\}_{i=1}^{M},\quad w_i \in \{w_{\mathrm{A1}}, w_{\mathrm{A2}}, w_{\mathrm{A3}}\},\\
\text{Negative rubrics: } &\mathcal{N} = \{n_j\}_{j=1}^{K},\quad \lambda_j \in \{\lambda_{\mathrm{S1}}, \lambda_{\mathrm{S2}}, \lambda_{\mathrm{S3}}, \lambda_{\mathrm{S4}}\},\\
\text{Indicators: } &h_i,g_j \in \{0,1\},\; h_i=1 \text{ if } p_i \text{ is satisfied},\; g_j=1 \text{ if } n_j \text{ is triggered.}
\end{aligned}
\]

Element-wise decisions $(h_i,g_j)$ are derived through a hybrid procedure:
\begin{itemize}
    \item For A1–A3 and S2–S4 elements, an ensemble of LLM judges (Section~\ref{subsubsec:llm-judge}) performs semantic and numeric matching against the rubric text.
    \item For S1 elements, a lightweight rule-based relevance matcher first compares the response to all positive rubric elements (A1–A3).  
    Segments that fail to align semantically with any rubric content—off-topic, redundant, or non-actionable phrases—are flagged as S1 and assigned small penalties.
\end{itemize}

The raw score aggregates credit and penalties:
\[
S_{\mathrm{raw}} = \sum_{i=1}^{M} w_i h_i - \sum_{j=1}^{K} \lambda_j g_j,
\qquad
S_{\mathrm{norm}} = \frac{S_{\mathrm{raw}}}{\sum_{i=1}^{M} w_i}.
\]
Normalized scores are clipped to \([0,1]\):
\[
S_{\mathrm{final}} = \min\!\bigl(1,\; \max\!\bigl(0,\; S_{\mathrm{norm}}\bigr)\bigr).
\]

Safety overrides are applied hierarchically.  
If any critical error (S4) is detected ($\exists j : g_j = 1 \wedge n_j \in \mathrm{S4}$), the score is immediately set to zero:
\[
S_{\mathrm{final}} = 0.
\]
Penalties for S1–S3 are finite and monotonic ($\lambda_{\mathrm{S1}} < \lambda_{\mathrm{S2}} < \lambda_{\mathrm{S3}}$), reflecting increasing clinical severity.

\subsubsection{LLM-as-a-Judge Ensemble}
\label{subsubsec:llm-judge}

Element-wise scoring is performed by an ensemble of large language models acting as impartial judges (Layer 2 in Figure~\ref{fig:evaluation-pipeline}), following frameworks such as HealthBench~\cite{arora2025healthbench}.  
This setup blends rubric-based reasoning for A/S elements with rule-based checks for S1 relevance.

\begin{itemize}
    \item \textbf{Ensemble composition:} Three independent LLMs from distinct architectures evaluate each rubric element. All runs use deterministic sampling (temperature $\leq 0.3$) without external retrieval to ensure reproducibility.
    \item \textbf{Positive rubric checks (A1–A3):} A hit is assigned only if the response semantically includes the rubric content, with correct qualifiers (e.g., stage, biomarkers) and numerical precision (units, thresholds).
    \item \textbf{Negative rubric checks (S2–S4):} A hit occurs when the response explicitly endorses, implies, or fails to refute an unsafe recommendation or contraindicated action.
    \item \textbf{Relevance rule (S1):} A supplementary rule-based module computes sentence-level alignment between the response and all positive rubric spans using semantic similarity and keyword overlap. Non-aligned text segments are marked as S1 violations and lightly penalized to encourage concise, focused reasoning without conflating irrelevance with clinical risk.
    \item \textbf{Majority voting:} Binary decisions $(h_i,g_j)$ from all judges are aggregated via majority vote, then combined through the weighting and normalization scheme above to yield $S_{\mathrm{final}} \in [0,1]$.
\end{itemize}

This hybrid rubric–rule scoring pipeline ensures that correctness, completeness, and safety are jointly quantified: the rubric framework captures clinical adequacy (A-axis) and harm risk (S2–S4), while the rule-based S1 module enforces relevance and brevity—together forming a unified, reproducible metric of clinical reasoning quality.

\subsubsection{Reliability Validation}
\label{subsubsec:scoring-validation}

To ensure the trustworthiness of the automated evaluation, we rigorously validated the framework against independent annotations from five senior clinical experts.
Across the GAPS benchmark \textit{GAPS-NSCLC-preview} of 92 questions and 1,691 individual rubrics, the automated pipeline achieved an overall agreement rate of 90.00\% with expert consensus. The Cohen's Kappa coefficient~\cite{cohen1960coefficient} was 0.77, indicating ``substantial agreement''.
These results confirm that the GAPS automated judge possesses expert-level reliability.

\section{Discussion}
\label{sec:discussion}

\subsection{Advantages of the GAPS Benchmark over Existing Evaluations}

Medical LLM benchmarks to date fall broadly into two paradigms:
(1) \textit{exam-style datasets} that test factual recall, such as MedQA, PubMedQA, MedMCQA; and  
(2) \textit{rubric-based evaluations} such as HealthBench that attempt to measure open-ended answers through human-designed scoring criteria.
Each line of work has contributed important foundations but remains limited in clinical validity and scalability.  
GAPS unifies the strengths of both paradigms while addressing their core limitations.

\subsubsection{Limitations of exam-style evaluations.}  
Traditional QA-style benchmarks treat medicine as a test of recall.  
They assess whether a model can reproduce the correct answer to a closed question, but not how it reasons or whether the reasoning is safe.  
These datasets flatten clinical cognition into a binary metric—right or wrong—ignoring the graded nature of medical expertise.  
As a result, models that memorize textbook facts can appear competitive with those capable of evidence-based reasoning, even though their clinical reliability differs substantially.  
Moreover, such datasets lack explicit modeling of uncertainty, patient context, and safety risk, making them poorly aligned with real-world medical decision-making.  
GAPS overcomes these shortcomings by defining an explicit hierarchy of reasoning depth (\textit{Grounding axis, G1–G4}) that mirrors the cognitive progression of clinical expertise—from factual recall and mechanistic explanation to applied and inferential reasoning under uncertainty.  
This design transforms evaluation from static knowledge testing into structured assessment of reasoning competence.

\subsubsection{Limitations of current rubric-based benchmarks.}  
HealthBench and similar frameworks have advanced the field by introducing rubric-guided evaluation, yet they remain constrained by manual design.  
Their rubrics lack an explicit medical ontology or evidence basis: each key point is written and weighted by humans without a transparent rationale for why one element is prioritized over another.  
Consequently, the evaluation lacks internal consistency—different items may be scored under divergent, subjective standards—and cannot scale across specialties.  
More importantly, HealthBench provides no formal link between its rubrics and the medical evidence hierarchy, leaving it unclear whether “correct” answers correspond to authoritative, guideline-supported actions.  
GAPS directly addresses these issues through a \textit{GRADE-based DeepResearch pipeline} that generates rubrics automatically from authoritative guidelines.  
Each positive or negative rubric element is derived from verifiable evidence within a bounded \textit{evidence neighborhood}, yielding a homogeneous, reproducible, and clinically interpretable scoring foundation.

\subsubsection{Unified multidimensional evaluation.}  
By integrating structured reasoning depth (G-axis), graded adequacy (A-axis), perturbation robustness (P-axis), and safety stratification (S-axis), GAPS moves beyond both paradigms.  
It evaluates not only whether an answer is correct but also whether it is complete, stable under noisy inputs, and safe in its implications.  
This multidimensional framework aligns evaluation with competency-based assessment in human medicine and provides a reproducible, evidence-centered foundation for diagnosing failure modes in AI clinician systems.

\subsection{Model Weaknesses and Insights on Future Development}

Evaluation across the GAPS axes shows that current large language models perform strongly at lower reasoning levels but struggle as cognitive complexity and input variability increase.  
At G1 (factual recall) and G2 (explanatory reasoning), all evaluated models achieved consistently high scores, indicating near-saturated capabilities in retrieving and articulating explicit guideline knowledge.  
However, when reasoning requires evidence application, uncertainty management, or robustness to misleading inputs, three structural limitations become apparent.

\subsubsection{G2–G3 Gap: From Abstract to Specific}  
Performance drops substantially between G2 (explanatory reasoning) and G3 (applied decision-making).  
This transition marks the shift from abstract interpretation to specific clinical judgment, representing the movement from conceptual understanding to practice-level competence. 
Models can memorize knowledge, and describe mechanisms, yet they often fail to apply those abstraction into concrete patient contexts.   
Even when all relevant information is present in the prompt, they mis-prioritize or omit key qualifiers, producing plausible but clinically invalid recommendations.  
This abstract-to-specific failure reflects the absence of procedural reasoning: LLMs operate at a propositional level but lack the ability to perform structured decision control.  
Bridging this gap will require systems capable of contextual grounding---integrating explicit guideline logic, decision paths, and constraint-aware inference to transform abstract knowledge into patient-specific, guideline-aligned recommendations.

\subsubsection{G3–G4 Gap: From Certainty to Uncertainty}  
A second and steeper decline occurs between G3 (evidence-based application) and G4 (inferential reasoning under uncertainty).  
When faced with incomplete or conflicting evidence, models display erratic behavior, oscillating between overconfident extrapolation and vague answers.  
They attempt analogical reasoning but rarely articulate uncertainty or define the boundaries of inference.  
This pattern reveals a fundamental limitation of current model cognition: reasoning by frequency rather than causality.  
Unlike clinicians, who reason by analogy while explicitly qualifying uncertainty, models tend to generalize beyond evidence without signaling confidence or reliability.  
Progress at this level will depend on the development of meta-reasoning mechanisms for causal abstraction, probabilistic reasoning, and structured uncertainty representation.

\subsubsection{Robustness Gap: Sensitivity to Premise Integrity}

The Perturbation axis further demonstrates that models remain brittle when input integrity is compromised.  
While they tolerate linguistic variation (P1) and redundant information (P2), performance collapses under P3 perturbations that introduce subtle misinformation or logical inversions.  
Most models fail to detect or correct a false premise, particularly when it contradicts a Must-have rubric element.  
This fragility suggests that models process clinical text descriptively rather than diagnostically; they track language coherence but not factual coherence.  
Improving robustness will require explicit mechanisms for internal consistency checking, counterfactual reasoning, and premise verification during inference.

\subsubsection{Towards Next-Generation AI Clinician Systems} 

These reasoning and robustness gaps define the next developmental frontier for AI clinician systems.  
Future research should focus on three complementary directions.

\begin{itemize}
    \item \textbf{Evidence-grounded reasoning.}  
    A clinical ready AI system must internalize not only medical content but also the hierarchical logic of evidence importance.  
    Embedding GRADE-like evidence-based-medicine methodologies into the reasoning process can help models weigh evidence strength and applicability instead of relying solely on textual proximity.

    \item \textbf{Procedural decision control.}  
    To close the G2–G3 gap, decision reasoning should be formalized as a structured process, integrating causal preconditions, contraindication checks, and temporal logic for action sequencing.  
    Such mechanisms could transform LLMs from descriptive narrators into operational decision agents.

    \item \textbf{Uncertainty- and premise-aware inference.}  
    For the G3–G4 and robustness gaps, models should learn to recognize when information is insufficient, contradictory, or unsafe.  
    Incorporating adversarial training on P3-style perturbations and uncertainty-aware objectives can promote cautious, self-corrective reasoning in ambiguous clinical contexts.
\end{itemize}

In summary, GAPS demonstrates that current large models have largely mastered factual and explanatory reasoning but remain limited in three critical areas: translating evidence into actions, reasoning beyond direct evidence, and maintaining reliability under misleading inputs.  
Addressing these challenges—from knowing to doing, from doing to reasoning under uncertainty, and from reasoning to verifying the premise—will be essential to advancing LLMs from encyclopedic assistants to trustworthy clinical reasoning partners.

\section{Conclusion}
\label{sec:conclusion}

This study introduces the GAPS framework and its fully automated benchmark pipeline for evaluating large medical language models across four clinically grounded dimensions: reasoning depth, answer adequacy, input robustness, and safety.  
By anchoring every evaluation item to authoritative evidence and by automating rubric synthesis under GRADE principles, GAPS provides a scalable and reproducible foundation for assessing AI clinicians not only on factual accuracy but on their ability to reason, decide, and act safely.  
Through systematic analysis, we show that while contemporary large language models excel at factual and explanatory reasoning (G1–G2), their performance deteriorates at higher cognitive levels and under perturbed or adversarial inputs, revealing persistent evidence-to-action and inferential reasoning gaps.  
These findings highlight the need for next-generation systems that integrate evidence-grounded reasoning, structured decision control, and uncertainty-aware inference to achieve true clinical reliability.

Despite these advances, several limitations remain.  
First, current GAPS items are formulated as single-turn questions rather than interactive dialogues, which constrains the evaluation of longitudinal reasoning, patient communication, and iterative decision-making.  
Second, all benchmark content is text-based; multimodal clinical information such as imaging, laboratory values, and temporal trends is not yet incorporated.  
Third, the current perturbation design, while systematic, represents only an early approximation of real-world variability and does not yet capture the full spectrum of human ambiguity, data noise, and contextual inconsistency encountered in clinical practice.  
Finally, although rubric generation is automated, human-in-the-loop validation remains essential to ensure semantic fidelity and clinical safety.  

Future work will extend GAPS toward multimodal, conversational, and continuously adaptive evaluation settings, integrating real patient trajectories and dynamic evidence updates.  
By progressively closing the gap between benchmark design and clinical reality, GAPS aims to serve not only as an assessment tool but as an evolving framework for shaping the safe, interpretable, and evidence-based development of AI clinician systems.

\section*{Data Availability}
The benchmark dataset constructed for this study (\textit{GAPS-NSCLC-preview}) is available for research purposes at \url{https://huggingface.co/datasets/AQ-MedAI/GAPS-NSCLC-preview}.

\section*{Code Availability}
The source code for the evaluation scripts is publicly available on GitHub at \url{https://github.com/AQ-MedAI/MedicalAiBenchEval}. The automated benchmark construction pipeline and the DeepResearch rubric agent will be made available at a later date.

% \begin{appendices}

% \section{Section title of first appendix}\label{secA1}

% An appendix contains supplementary information that is not an essential part of the text itself but which may be helpful in providing a more comprehensive understanding of the research problem or it is information that is too cumbersome to be included in the body of the paper.




% \end{appendices}


\bibliography{main-bibliography}% common bib file

\end{document}